\documentclass[runningheads]{llncs}
\usepackage{graphicx}
\usepackage{subfig}
\usepackage{amssymb}
\usepackage{amsmath} 
\usepackage{tabularx}
\usepackage{booktabs}

\begin{document}
\title{A Dense CNN approach for skin lesion classification}
%
%
\author{Pierluigi Carcagn\`i\inst{1} \and
Andrea Cuna\inst{2} \and
Cosimo Distante\inst{1}}
\authorrunning{P.Carcagni et al.}
%
\institute{CNR-ISASI. Ecotekne Campus via Monteroni snc, 73100 Lecce, Italy \and
University of Salento. Ecotekne Campus via Monteroni snc, 73100 Lecce, Italy\\
\email{p.carcagni@isasi.cnr.it}\\
}
\maketitle              
\begin{abstract}
This article presents a Deep CNN, based on the DenseNet architecture jointly with a highly discriminating learning methodology, in order to classify seven kinds of skin lesions: Melanoma, Melanocytic nevus, Basal cell carcinoma, Actinic keratosis / Bowen's disease, Benign keratosis, Dermatofibroma, Vascular lesion. In particular a  61 layers DenseNet, pre-trained on IMAGENET dataset, has been fine-tuned on ISIC 2018 Task 3 Challenge Dataset exploiting a Center Loss function. 

\end{abstract}
\section{Introduction}
Convolutional Neural Networks (CNN) are increasingly used in the computer vision field for tasks such as classification, detection and regression. In this paper is reported a CNN based approach to tackle \textit{Task 3: Lesion Diagnosis of the ISIC 2018: Challenge Skin Lesion Analysis Towards Melanoma} \cite{ISIC2018}. This task is related to the automated predictions of disease classification within dermoscopic images. The possible disease categories are: Melanoma, Melanocytic nevus, Basal cell carcinoma, Actinic keratosis / Bowen's disease (intraepithelial carcinoma), Benign keratosis (solar lentigo / seborrheic keratosis / lichen planus-like keratosis), Dermatofibroma, Vascular lesion. 
\\

In particular, a CNN architecture, derived from the DenseNet one proposed in \cite{huang2017densely}, has been employed  in order to classify the demoscopic lesion images   of  the seven categories indicated in the challenge. Moreover, during the training phase, a linear combination of \textit{soft-max}  and \textit{center-loss} \cite{wen2016discriminative} functions are employed as total loss function. 
In Section 2, the proposed classification system is described and in Section 3 the experimental results. Finally,  Section 4 reports conclusions and  future works. 


\section{Methodology}
The proposed classification system is based on the  DenseNet-121  architecture described in \cite{huang2017densely}. In particular, the
pretrained, by means of IMAGENET dataset \cite{deng2009imagenet}, version has been exploited considering a reduced verion (61 layers)  and fine-tuned on \textit{ISIC 2018 Challeng Task 3 Training Dataset}. This dataset data was extracted from the \textit{ISIC 2018: Skin Lesion Analysis Towards Melanoma Detection} grand challenge datasets \cite{codella2018skin}\cite{tschandl2018ham10000}.
In order to obtain a high discriminative CNN model, in the learning phase the center-loss function based approach proposed in \cite{wen2016discriminative} has been exploited. In particular, in the course of CNN training, high discriminative features are learned considering jointly softmax and center loss functions balanced by means of a hyper parameter. In \cite{wen2016discriminative} the center loss function is defined by:

\begin{equation}\label{equ:centeloss}
L_{center} = \frac{1}{2} \sum_{i=1}^{m} \lVert \bold{x}_i - \bold{c}_{y_i} \rVert_2^2
\end{equation}

where the term  $\bold{c}_i\in\Re^d$ denotes the $y_i$th class center of 
deep features $\bold{x}_i$.
\\\\
Finally, the total loss function is defined as linear combination of soft-max  $L_s$  and center-loss $L_c$ functions as following:

\begin{equation}\label{equ:total_loss}
L=L_s+\lambda{L_c}
\end{equation}

where the term $\lambda$ is a scalar used for balancing the two loss functions. Intra-class minimizations, during the learning phase, are controlled by means of $L_c$, inter-class maximizations by means of $L_s$.
\\\\
%

\section{Experimental Results}

The training dataset used in this work is that provided by the ISIC for the 2018 challenge, consisting of approximately 10015 images covering 7 types of skin lesions as reported in Tab.~\ref{tab::dataset}.

\begin{table}
\centering
\caption{ISIC 2018 Challeng Task 3 Training Dataset. Number of images for each class.}\label{tab::dataset}
\begin{tabular}{|l|l|l|l|l|l|l|l|}
\hline
\textbf{MEL} &  \textbf{NV} & \textbf{BCC} & \textbf{AKIEC} & \textbf{BKL} & \textbf{DF} & \textbf{VASC} & \textbf{Total Number of Images}\\
\hline
1113 & 6705 & 514 & 327 &1099 & 115 & 142 & 10015\\
\hline
\end{tabular}
\end{table}

The CNN used  is composed as follows.
Starting from DenseNet-121, we modified it in such a way as to keep the same structure inside each dense block, but by reducing their number from 4 to 3.
The network has been "cut" near the \textit{concat\_4 \_11} layer (in the third dense block). The network thus becomes composed of 3 dense layers, where the third layer is composed of 22 convolutional layers (instead of 48 as in DenseNet-121).

We used a DenseNet-BC structure as described in \cite{huang2017densely}, but with 3 dense blocks on $224\times224$ input images. 
As transition layers between two contiguous dense blocks, we used $1\times1$ convolution followed by $2\times2$ average pooling. At the end of the last dense block, a global average pooling is performed (\textit{pool5}) before the final output layer.

Finally, in the fine-tuning phase, layers from the \textit{concat\_4\_6}  to the fully connected one (last layer) were trained from scratch keeping blocked all the previous layers.

Given the complexity of the network and the number of parameters to be trained during the fine-tuning procedure, it was necessary to increase the samples of the starting dataset. The procedure used was as follows. In a first phase, the original dataset was divided into two parts: one for training and the other for validation during network training. In particular, for each class 80\% of images were considered for training and the remaining 20\% for validation. The subdivision was made so that there were no common images between the training and validation sets for each class of images.
\\
Then we proceeded to increase the number of images through geometric transformations such as rotations, flipping and affine 
in order to obtain a dataset that is as balanced as possible as reported in Tab.~\ref{tab::splitteddataset}. Downstream of the geometrical transformations, a centered and square cropping was made, of amplitude equal to the shorter side of the starting image. Thus, the resulting patch was resized to a dimension of $224\times224$ pixels as requested by the input layer of the network.


Network training was performed using two NVIDIA GTX 1080Ti cards and the Caffe  \cite{jia2014caffe} framework. As optimizer, SGD was chosen with learning rate starting at 0.01, weight decay and momentum equal to 0.0001 and 0.9 respectively.
The maximum number of iterations has been set at 75000, decreasing the learning rate by a factor of 10 at each step of 20000 iterations. Finally, the 0.8 value was used for the $\lambda$ parameter in the Eq. ~\ref{equ:total_loss}

\begin{table}
\centering
\caption{Dataset split: 80\% training; 20\% test.}\label{tab::splitteddataset}
\begin{tabular}{|l|l|l|l|l|l|l|l|l|}
\hline
&\textbf{MEL} &  \textbf{NV} & \textbf{BCC} & \textbf{AKIEC} & \textbf{BKL} & \textbf{DF} & \textbf{VASC} & \textbf{Total}\\
\hline
\textbf{Augmented Dataset}& 1113 & 6705 & 514 & 327 &1099 & 115 & 142 & 10015\\
\hline
\textbf{Augmented Training Dataset}& 891 & 5364 & 412 & 262 & 880 & 92 & 114 & 8015\\
\hline
\textbf{Augmented Test Dataset}& 222 & 1341 & 102 & 65 & 219 & 23 & 28 & 2000\\
\hline
\end{tabular}
\end{table}

\begin{table}
\centering
\caption{Balanced  Dataset}\label{tab::augmentedataset}
\begin{tabular}{|l|l|l|l|l|l|l|l|l|}
\hline
&\textbf{MEL} &  \textbf{NV} & \textbf{BCC} & \textbf{AKIEC} & \textbf{BKL} & \textbf{DF} & \textbf{VASC} & \textbf{Total}\\
\hline
\textbf{Augmented Dataset}& 50529 & 87165 & 46774 & 45453 & 51653 & 43815 & 43990 & 369379\\
\hline
\textbf{Augmented Training Dataset}& 40095 & 69732 & 37492 & 36418 &41360 & 35052 & 35226 & 295375\\
\hline
\textbf{Augmented Test Dataset}& 10434 & 17433 & 9282 & 9035 &10293 & 8763 & 8764 & 74004\\
\hline
\end{tabular}
\end{table}

The proposed system performed an overall score of 89.2 \% in the validation phase of the challenge.

\section{Conclusions and Future Work}

In this work, a multi-class approach for skin lesion classification, based on deep CNNs, has been proposed. The system doesn't  require any particular pre-porocessing of the input images and obtaining high accuracy rate in the ISIC 2018 Task 3 Validation phase.
\\
We expect improved performances if a more robust pre-processing step is employed before to supply images in input to the proposed system, such as the use of segmentation in order to obtain registered images into a common reference.

 \bibliographystyle{splncs04}
 \bibliography{references}
%
%
%
%
%
\end{document}